\documentclass[conference]{IEEEtran}
\IEEEoverridecommandlockouts
\usepackage{cite}
\usepackage{amsmath,amssymb,amsfonts}
\usepackage{algorithmic}
\usepackage{graphicx,color}
\usepackage{siunitx}
\usepackage{textcomp}
\usepackage{amsmath,amsfonts}
\usepackage{algorithmic}
\usepackage{algorithm}
\usepackage{array}
\usepackage{textcomp}
\usepackage{stfloats}
\usepackage{url}
\usepackage{verbatim}
\usepackage{graphicx}
\usepackage{cite}
\DeclareMathAlphabet\mathbfcal{OMS}{cmsy}{b}{n}
\usepackage{algorithm}      
\usepackage{algorithmic}

\usepackage{varwidth}
\usepackage{float}
\usepackage{graphicx}
\usepackage{subcaption}
\usepackage{balance}
\usepackage{xcolor}
\usepackage{booktabs}
\usepackage{hyperref}
\usepackage{array, multirow}
\usepackage{float}
\usepackage{adjustbox}
\usepackage{bbm}
\usepackage{url}
\def\BibTeX{{\rm B\kern-.05em{\sc i\kern-.025em b}\kern-.08em
    T\kern-.1667em\lower.7ex\hbox{E}\kern-.125emX}}
    
\usepackage{textcomp}
\usepackage{amsmath,amssymb,amsfonts}

\def\BibTeX{{\rm B\kern-.05em{\sc i\kern-.025em b}\kern-.08em
    T\kern-.1667em\lower.7ex\hbox{E}\kern-.125emX}}
\begin{document}

\title{Context-aware Attention-based Gaussian Mixture Models for Vehicular Trajectory Prediction}
\author{\IEEEauthorblockN{Arash Raftari\IEEEauthorrefmark{1}, Babak Ebrahimi Soorchaei\IEEEauthorrefmark{2}, Yaser P. Fallah\IEEEauthorrefmark{1}}
\IEEEauthorblockA{\IEEEauthorrefmark{1}Department of Electrical and Computer Engineering, University of Central Florida, Orlando, Florida, USA\\
\{arash.raftari, yaser.fallah\}@ucf.edu}
\IEEEauthorblockA{\IEEEauthorrefmark{2} Department of Computer Science, University of Central Florida, Orlando, Florida, USA\\
 babak.ebrahimi.soorchaei@ucf.edu} 

 \thanks{This work was supported in part by a donation from AMD to the University of Central Florida Department of Electrical and Computer Engineering's robotics initiatives.}
 \thanks{\textcopyright~2026 IEEE. Personal use of this material is permitted. Permission from IEEE must be obtained for all other uses, in any current or future media, including reprinting/republishing this material for advertising or promotional purposes, creating new collective works, for resale or redistribution to servers or lists, or reuse of any copyrighted component of this work in other works.}}

\maketitle

\begin{abstract}
Reliable and interpretable trajectory prediction is critical for cooperative and autonomous driving in complex and uncertain environments. This paper introduces a Context-Aware Attention-based Gaussian Mixture Model (CAA-GMM) for multimodal, uncertainty-aware motion forecasting.
The proposed approach models future motion as a probabilistic mixture conditioned on both scene context and agent dynamics, capturing diverse behavioral modes with interpretable Gaussian components. A lightweight attention mechanism adaptively encodes inter-agent interactions and contextual salience, enabling efficient fusion of rasterized environment cues and motion history in dense traffic scenes.
Comprehensive evaluations on the nuScenes and Argoverse 2 datasets demonstrate that CAA-GMM achieves competitive or superior accuracy compared with state-of-the-art raster-based baselines, while maintaining low computational complexity. Ablation analyses confirm the importance of the attention module for robust contextual reasoning and predictive precision. Furthermore, evaluations under imperfect communication and perception conditions highlight the framework’s resilience to uncertainty, establishing CAA-GMM as an efficient and scalable solution for cooperative trajectory prediction in intelligent transportation systems.
\end{abstract}

\begin{IEEEkeywords}
Trajectory Prediction, Autonomous Driving, Connected and Autonomous Vehicles (CAVs), Vehicle-to-Everything (V2X) Communication, Attention Mechanism, Gaussian Mixture Model (GMM), Probabilistic Forecasting
\end{IEEEkeywords}

\section{INTRODUCTION}

Accurate trajectory prediction is a cornerstone of autonomous and connected driving, enabling safe planning and cooperative decision-making in dynamic traffic environments. Vehicles must anticipate the future motions of surrounding agents—including vehicles, cyclists, and pedestrians—to safely navigate complex scenarios such as intersections, merges, and dense urban traffic~\cite{lefevre2014survey,rasouli2020autonomous}. This task is inherently challenging due to the multimodal nature of human behavior, sensor noise, occlusions, and uncertainty arising from partial observability and interaction dynamics~\cite{salzmann2020trajectronpp,chai2019multipath,deo2018convolutional,cui2019multimodal}.

Recent deep learning approaches for trajectory prediction are commonly categorized into raster-based, graph-based, and planning-aware paradigms. Raster-based methods encode agents and map context as BEV images processed by convolutional or transformer architectures~\cite{bansal2018chauffeurnet,phan2020covernet,varadarajan2022multipathpp}, with recent attention-based variants such as CASPFormer improving contextual reasoning while remaining computationally efficient~\cite{wu2024caspformer}

Graph-based approaches explicitly model agents and map elements as nodes connected via relational edges, capturing interactions through message passing or self-attention~\cite{gupta2018social,gao2020vectornet,liang2020learning,ngiam2022wayformer,zhou2022hivt}. These methods excel at interaction modeling but often require high-definition maps and complex preprocessing pipelines, limiting scalability and robustness in partially observed or communication-constrained settings. Planning-aware models further integrate intent or goal inference to couple prediction and decision-making~\cite{ivanovic2019trajectron,mangalam2020pecnet,sun2024lapred}. Although effective in structured autonomous scenarios, their reliance on rational intent assumptions and explicit goal annotations can reduce performance in mixed or human-driven traffic.

In parallel, cooperative perception and prediction have gained increasing attention with the emergence of Vehicle-to-Everything (V2X) communication. Model-Based Communication (MBC) frameworks propose exchanging compact probabilistic representations—rather than dense trajectories or raw sensor data—to reduce bandwidth while preserving predictive fidelity~\cite{fallah2018mbc,kim2020model,kim2021cooperative,10189177}. However, existing learning-based predictors are not explicitly designed to operate under realistic communication constraints, nor do they jointly account for communication loss and perception degradation.

To address these limitations, this paper proposes the \textbf{Context-Aware Attention-based Gaussian Mixture Model (CAA-GMM)}, a lightweight probabilistic trajectory prediction framework tailored for cooperative driving. CAA-GMM operates within the MBC paradigm by transmitting Gaussian mixture parameters, achieving communication efficiency while remaining compatible with SAE J2735 Basic Safety Message constraints~\cite{3gpp23285_archit}. The model fuses rasterized static-map features with learned embeddings of agent motion histories and neighboring interactions using a transformer-based attention mechanism. A Gaussian mixture decoder then produces interpretable multimodal trajectory distributions with explicit uncertainty quantification.

Comprehensive evaluations on \textbf{nuScenes}~\cite{caesar2020nuscenes} and \textbf{Argoverse 2}~\cite{wilson2021argoverse2} demonstrate that CAA-GMM achieves competitive or superior accuracy compared to state-of-the-art raster-based methods while significantly reducing computational and communication overhead. Ablation studies confirm the critical role of attention in modeling dynamic context, and experiments incorporating empirical V2X channel and observation-loss models show robust performance under packet loss and degraded perception. To the best of our knowledge, this is the first work to explicitly evaluate the joint impact of communication and observation losses on learning-based trajectory prediction.

\section{Context-Aware Attention-based Gaussian Mixture Models}
\label{sec:CAAGMM_full}

This section presents the proposed \emph{Context-Aware Attention-based Gaussian Mixture Model (CAA-GMM)} for vehicular trajectory prediction. We first briefly introduce a reduced variant, the \emph{Context-Aware Gaussian Mixture Model (CA-GMM)}, which provides the underlying Bayesian formulation for multimodal inference. CA-GMM is described only to establish notation, mixture-based conditioning, regression, and mode construction. We then present CAA-GMM, our main contribution, which extends CA-GMM with mode-level and interaction-level attention to enable socially consistent and context-sensitive prediction.

\subsection{Context-Aware Gaussian Mixture Model (CA-GMM)}
\label{sec:CAGMM}

We briefly describe the Context-Aware Gaussian Mixture Model (CA-GMM), which provides the probabilistic foundation and notation for the proposed CAA-GMM.

\begin{figure*}[!t]
\centering
\includegraphics[width=.9\textwidth]{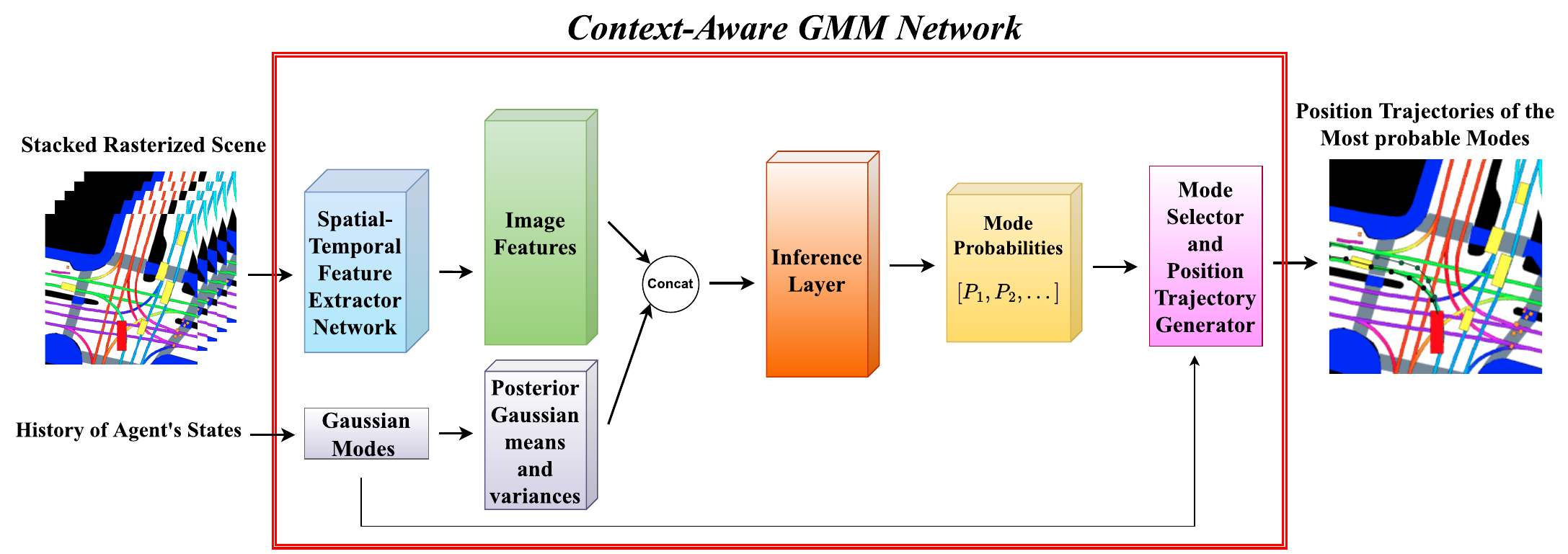}
\caption{Context-aware GMM architecture (CA-GMM), shown as a reduced variant of CAA-GMM for ablation and formulation clarity.}
\label{CGMM}
\vspace{-0.35cm}
\end{figure*}

\subsubsection{Gaussian Mixture Modeling and Mode Inference}
Let the $d$-dimensional motion vector $\mathbf{X}$ (e.g., longitudinal speed or heading over the prediction horizon) be modeled as a $K$-component Gaussian mixture:
\begin{equation}
P(\mathbf{X}) = \sum_{k=1}^{K} \pi_k \mathcal{N}(\mathbf{X} \mid \mu_k, \Sigma_k).
\end{equation}
Each component represents a distinct motion mode.

Given observed history $\mathcal{V}^{obs}$ and future variables $\mathcal{V}^{\ast}$, the joint distribution is
\begin{equation}
\begin{bmatrix}\mathcal{V}^{obs}\\\mathcal{V}^{\ast}\end{bmatrix}
\sim \sum_{k=1}^{K} \pi_k\,
\mathcal{N}\!\left(
\begin{bmatrix}\mathcal{V}^{obs}\\\mathcal{V}^{\ast}\end{bmatrix}
\Big| \mu_k, \Sigma_k
\right),
\end{equation}
where $\mu_k$ and $\Sigma_k$ are partitioned into observed and future blocks.

Conditioned on $\mathcal{V}^{obs}$, the posterior likelihood of mode $k$ is
\begin{equation}
\mathbb{P}(k \mid \mathcal{V}^{obs}) \propto
\pi_k\,\mathcal{N}(\mathcal{V}^{obs} \mid \mu_k^{r}, \Sigma_{k11}),
\end{equation}
and the corresponding predictive distribution follows from Gaussian conditioning:
\begin{align}
\mu_k^{\ast} &= \mu_k^{d-r} + \Sigma_{k21}\Sigma_{k11}^{-1}(\mathcal{V}^{obs}-\mu_k^{r}),\\
\Sigma_k^{\ast} &= \Sigma_{k22} - \Sigma_{k21}\Sigma_{k11}^{-1}\Sigma_{k12}.
\end{align}

\subsubsection{Adaptive EM and Mode Construction}
Gaussian components are learned using an adaptive Expectation–Maximization (EM) procedure initialized via collapsed Gibbs sampling with conjugate priors. To ensure bounded predictive uncertainty, each component is evaluated based on the posterior covariance of the future variables,
\(
\Sigma_{\text{post}} = \Sigma_{22} - \Sigma_{21}\Sigma_{11}^{-1}\Sigma_{12}.
\)
Components whose maximum posterior variance exceeds a threshold $\tau$ are recursively split and reclustered until all modes satisfy the variance constraint.

In this work, separate GMMs are learned for longitudinal speed and heading. We consider
\(
\max \mathrm{diag}(\Sigma^{speed}_i) < 7 \text{ or } 10
\)
and
\(
\max \mathrm{diag}(\Sigma^{heading}_i) < 0.0514 \text{ or } 0.0685,
\)
corresponding to maximum standard deviations of $2.64$–$3.16$ m/s and $13^\circ$–$15^\circ$, respectively. These thresholds control the number of modes and balance expressiveness with numerical stability.

\subsubsection{Context-Aware Mode Classification}
To incorporate scene context, a neural classifier maps contextual features to a probability vector
$P(\text{context}) = [p_1,\dots,p_K]$,
which reweights the mixture posterior:
\begin{equation}
\mathbb{P}(k \mid \mathcal{V}^{obs}, \text{Context}) \propto
p_k\,\pi_k\,\mathcal{N}(\mathcal{V}^{obs} \mid \mu_k^{r}, \Sigma_{k11}).
\end{equation}
\begin{figure*}[!t]
\centerline{\includegraphics[width=.95\textwidth]{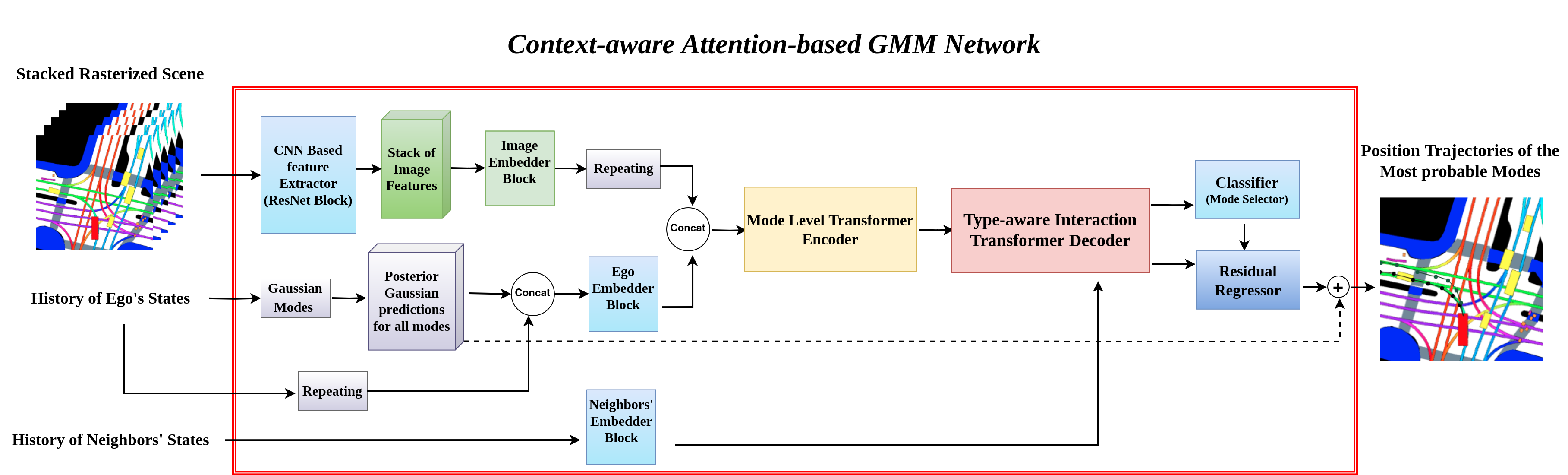}}
\caption{Context-Aware Attention-based GMM Architecture (CAA-GMM).}
\label{fig:CAAGMM}
\end{figure*}
Context is encoded using rasterized BEV representations of map elements and surrounding agents processed by a CNN backbone, while the Gaussian posterior parameters are concatenated with learned features to infer mode probabilities.

\subsubsection{Trajectory Reconstruction}
For each selected mode, predicted speed and heading distributions are integrated to obtain position using a Bayes Least Squares formulation, approximated in discrete time as
\begin{align}
\bar{x}_{j+1} &= \bar{x}_j + \mu_{s_j} e^{-\sigma_{h_j}^2/2} \cos(\mu_{h_j}),\\
\bar{y}_{j+1} &= \bar{y}_j + \mu_{s_j} e^{-\sigma_{h_j}^2/2} \sin(\mu_{h_j}).
\end{align}

\subsubsection{Training Objective}
\label{sec::training}
During training, each trajectory is assigned to its closest Gaussian mode based on $\ell^2$ distance in speed and heading space. The model is supervised using a combination of (i) categorical cross-entropy to encourage confident mode selection,
\begin{equation}
\mathcal{L}_{\text{entropy}} = -\sum_i y_i \log p_i - \sum_i z_i \log q_i,
\end{equation}
(ii) a mode-based regression loss enforcing spatial accuracy of the most confident prediction,
\begin{equation}
\mathcal{L}_{\text{mode\_error}} = \|\hat{x}_{\max} - x_{\text{gt}}\|_2,
\end{equation}
and (iii) a minimum average displacement error (minADE) loss to ensure multimodal coverage:
\begin{equation}
\mathcal{L}_{\text{min\_ade}} =
\min_{k \in \text{top-}K}
\frac{1}{t_{fut}} \sum_t \|\hat{x}^{(k)}_t - x^{gt}_t\|_2.
\end{equation}
The final loss is a weighted sum of these terms, with $\lambda_1=0.2$, $\lambda_2=0.3$, and $\lambda_3=0.5$, and $\text{top-}K=10$ during training and inference.
\begin{equation}
\label{loss}
\mathcal{L} =
\lambda_1 \mathcal{L}_{\text{entropy}} +
\lambda_2 \mathcal{L}_{\text{mode\_error}} +
\lambda_3 \mathcal{L}_{\text{min\_ade}}
\end{equation}

\subsection{Context-Aware Attention-based Gaussian Mixture Model (CAA-GMM)}
\label{sec:chap3_CAAGMM}
Accurate multi-agent trajectory prediction requires modeling both contextual cues and inter-agent dependencies. While the Context-Aware Gaussian Mixture Model (CA-GMM) provides a strong probabilistic baseline by combining Bayesian inference with contextual classification, it models each agent independently and lacks explicit interaction reasoning.  
To overcome this limitation, we propose the \textbf{Context-Aware Attention-based Gaussian Mixture Model (CAA-GMM)}, an extension of CA-GMM that incorporates transformer-based attention mechanisms to model dynamic, type-aware interactions between agents.  
By combining raster-based visual context with structured agent states (positions, velocities, and headings), CAA-GMM learns to weigh the influence of neighboring agents on the ego agent’s future motion, improving both accuracy and social consistency.

Inspired by recent lightweight transformer architectures for trajectory forecasting~\cite{shi2023tutr, xu2024positionfree, huang2025trajectorymamba, prutsch2024efficient, tang2024hpnet}, our model removes absolute and relative positional embeddings to enhance generalization across diverse layouts.  
In addition, we introduce a type-aware attention masking mechanism that modulates attention scores based on agent-type pairs (e.g., vehicle-to-vehicle, vehicle-to-pedestrian).  
This design allows the model to learn distinct interaction patterns between heterogeneous agents without requiring explicit type embeddings, resulting in a scalable and semantically structured formulation.

\subsubsection{System Architecture Overview}
\label{sec:chap3_CAAGMM_arch}

From a system-level perspective, CAA-GMM maintains the multi-modal probabilistic foundation of CA-GMM while introducing two additional transformer modules:  
(i) a \textbf{mode-level encoder} and  
(ii) a \textbf{type-aware interaction decoder}.  
Together, these modules enable both inter-modal reasoning (relationships among Gaussian trajectory modes) and inter-agent reasoning (social dependencies between agents).

\paragraph*{Mode-Level Transformer Encoder.}
The mode-level encoder captures dependencies across the $K$ Gaussian trajectory modes that would otherwise be treated independently.  
Each mode is represented as a token in the transformer sequence.  
At each encoder layer, the model applies \textbf{self-attention among mode tokens} to capture inter-modal dependencies (e.g., mutual exclusivity between turning and going straight), followed by \textbf{cross-attention between mode tokens and image embeddings} extracted from the BEV encoder.  
This design allows each mode to jointly reason about other plausible futures while grounding its representation in spatial scene context.  
Consistent with efficient transformer design~\cite{shi2023tutr, xu2024positionfree, huang2025trajectorymamba}, positional embeddings are omitted to reduce complexity and improve generalization.

\paragraph*{Type-Aware Interaction Decoder.}
To incorporate social reasoning, we introduce a transformer decoder that attends to neighboring agents based on spatial context and semantic type.  
Let the query matrix \(Q \in \mathbb{R}^{K \times d}\) represent the $K$ Gaussian modes from the encoder.  
Neighboring agents are grouped by type \(t \in \{v, p, b\}\), corresponding to vehicles, pedestrians, and bicyclists.  
Each group provides key and value matrices \(K^{(t)}, V^{(t)} \in \mathbb{R}^{N^{(t)} \times d}\), where \(N^{(t)}\) is the number of agents of type \(t\).  
For each type, we apply type-specific linear projections:
\[
Q^{(t)} = QW_Q^{(t)}, \quad K^{(t)} = K^{(t)}W_K^{(t)}, \quad V^{(t)} = V^{(t)}W_V^{(t)},
\]
with learned weights \(W_Q^{(t)}, W_K^{(t)}, W_V^{(t)} \in \mathbb{R}^{d \times d_h}\).
The \textbf{scaled dot-product attention} is then computed as
\[
A^{(t)} = \frac{Q^{(t)} {K^{(t)}}^\top}{\sqrt{d_h}},
\]
and modulated using a \textbf{type-specific attention mask} \(\mathcal{M}^{(t)} \in \{0, -\infty\}^{K \times N^{(t)}}\) to suppress occluded or irrelevant agents:
\[
\tilde{A}^{(t)} = \mathrm{Softmax}(A^{(t)} + \mathcal{M}^{(t)}), \quad 
\text{Attn}^{(t)} = \tilde{A}^{(t)}V^{(t)}.
\]
The final socially-conditioned representation aggregates across types:
\[
\text{Attn}_{\text{final}} = \sum_{t \in \{v, p, b\}} \text{Attn}^{(t)}.
\]
This formulation allows each Gaussian mode to selectively attend to relevant neighbors based on spatial influence and agent semantics, without increasing model complexity.
\begin{figure*}[!t]%
\centering
\subcaptionbox{Example of Nusenes HD map with its layers{\label{fig:hdmap}}}{%
\includegraphics[width=0.29\textwidth,trim={0mm 0mm 0mm 0mm},clip]{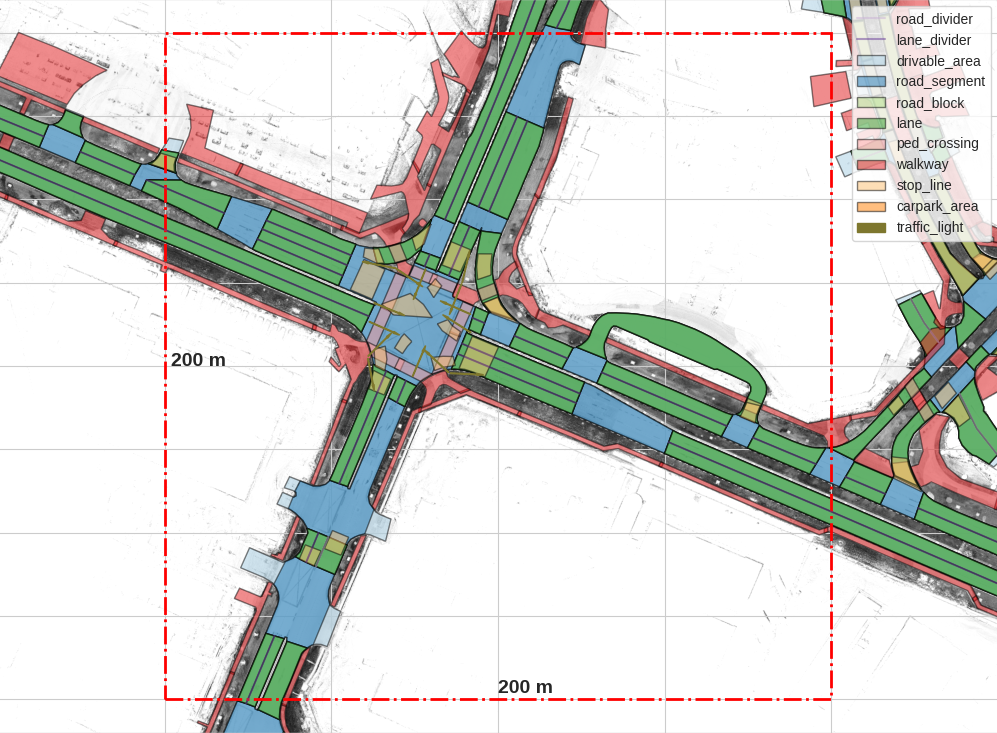}}%
\hspace{0.08in}
\subcaptionbox{Nuscenes raster image {\label{fig:raster_nu}}}{%
\includegraphics[width=0.22\textwidth,trim={0mm 0mm 0mm 0mm},clip]{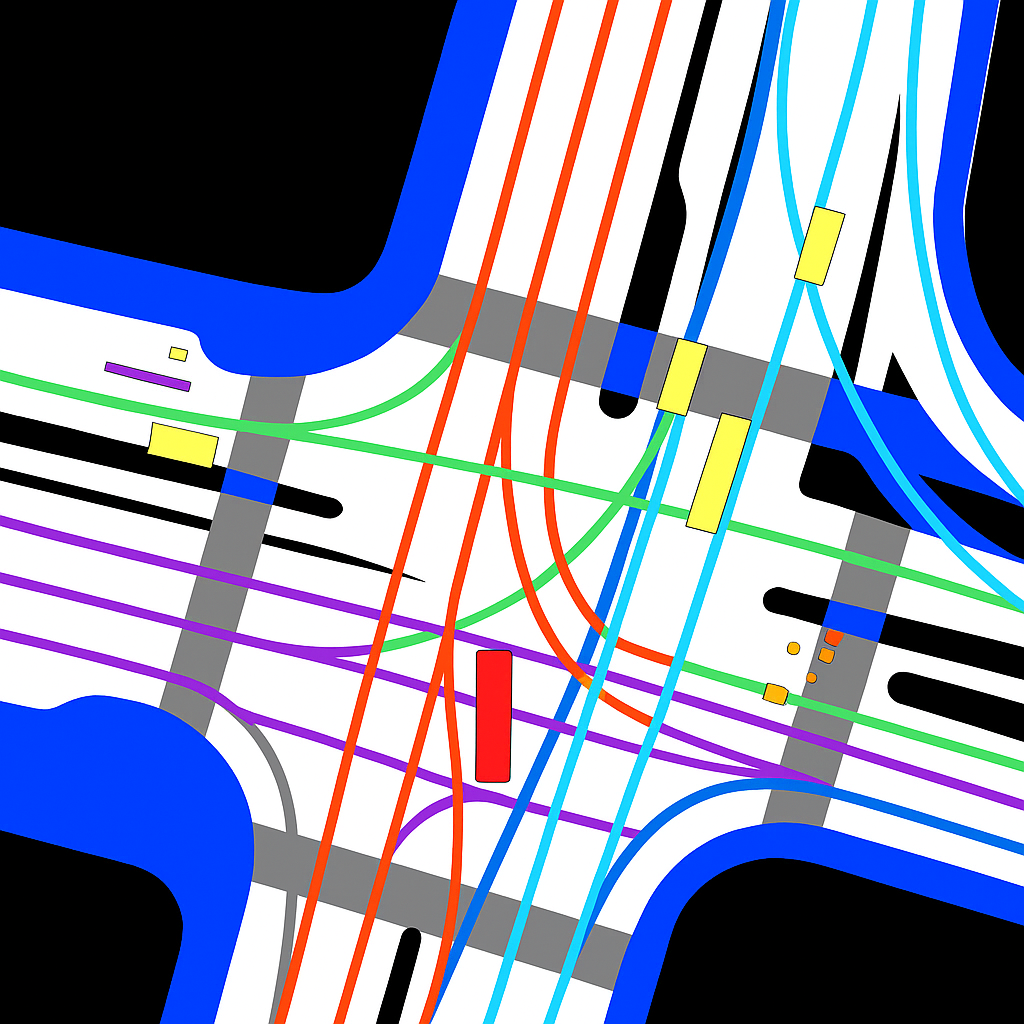}}%
\hspace{0.08in}
\subcaptionbox{AV2 raster image {\label{fig:raster_av}}}{%
\includegraphics[width=0.22\textwidth,trim={0mm 0mm 0mm 0mm},clip]{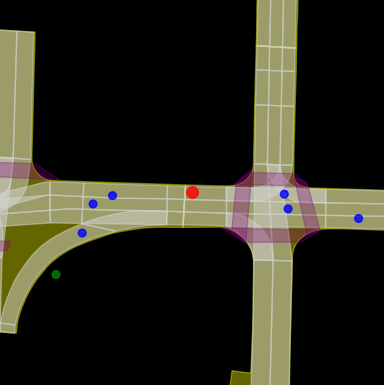}}%
\caption{HD map and BEV rasterized images}
\label{fig:dataset_nuscenes}
\end{figure*}
\subsubsection{Feature Extractors and Embedding Blocks}

Feature extraction from the rasterized BEV context is performed using a modified ResNet-34 backbone.  
The network is initialized with ImageNet-1k pretrained weights, employs an output stride of 8 via dilated convolutions, and integrates a feature pyramid network (FPN) for multi-scale fusion.  
The initial $7{\times}7$ convolutional stem is replaced with stacked $3{\times}3$ layers for improved spatial resolution, and normalization layers are frozen during warm-up to stabilize training.

The high-dimensional BEV features are processed by an \textbf{image embedder block} (Linear–ReLU–Linear).  
In parallel, the \textbf{ego embedder} encodes the concatenation of the ego state history and the posterior parameters (posterior mean for speed and heading and BLS estimation for position) of each Gaussian mode through a single linear layer.  
The output dimensions of both image and ego embeddings are matched so that they can be concatenated and fed into the mode-level encoder.

To represent other agents, a \textbf{neighbor embedder} projects their kinematic features (position, velocity, heading) via a single linear layer with an output size twice that of the ego embedding, allowing for richer representation of social interactions.  
Both transformer encoder and decoder blocks follow standard practice, comprising multi-head attention, layer normalization with dropout, and feedforward sublayers.

\subsubsection{Integration with the Bayesian Decoder and Training}

The interaction-aware mode representations are first scored by the classifier to produce logits over the $K$ Gaussian modes.  
The \textbf{top-$K$ modes} with the highest classifier confidence are selected, and their corresponding latent representations are forwarded to a lightweight \textbf{residual regressor}.  

The residual regressor is implemented as a simple \emph{MLP-Mixer–style} module that mixes information across \textbf{mode channels} and \textbf{temporal dimensions}.  
Given the selected mode embeddings, it predicts a residual trajectory vector of size \(T_{\text{fut}} \times 2\) (corresponding to $(\Delta x, \Delta y)$) for each selected mode.  
These residuals refine the Bayesian trajectory estimates produced by the GMM decoder, compensating for interaction-induced deviations that are not captured by the base Gaussian regression.

Final position trajectories are obtained by combining the Bayesian BLS estimation with the predicted residual offsets.  
CAA-GMM is trained end-to-end using the same multi-term objective~\eqref{loss} as CA-GMM, enabling consistent ablation and fair comparison.

\subsection{Reduced Variant (for Ablation): CA-GMM}
\label{sec:CAGMM_ablation}
For ablation, we define \textbf{CA-GMM} as the reduced form of CAA-GMM obtained by removing both the mode-level encoder and the type-aware interaction decoder.  
All remaining components—including the BEV encoder, Bayesian inference, and training objective—are kept identical.  
This ensures that performance differences isolate the effect of attention-based interaction modeling.

\section{Analysis and Results}
\label{sec:chap3_analysis}
We evaluate the proposed models on the \textbf{nuScenes}~\cite{caesar2020nuscenes} and \textbf{Argoverse~2}~\cite{wilson2021argoverse2} motion forecasting benchmarks. Both datasets provide large-scale multi-agent trajectories with HD map context across diverse urban environments. A unified preprocessing pipeline is applied to generate ego-centric bird’s-eye-view (BEV) raster representations aligned with the ego heading.
\subsection{Datasets}
\label{sec:datasets}

\textbf{nuScenes} contains 1,000 urban driving scenes. Following the standard protocol, we predict 6\,s of future motion given 2\,s of history, yielding over 34,000 annotated trajectories. Rasterized BEV inputs encode static map elements together with dynamic agents.

\textbf{Argoverse~2 (AV2)} provides large-scale multi-agent trajectories with complex lane topology. Models are trained using 5\,s of history for benchmark evaluation and 3\,s for ablation studies under communication and observation loss.

\subsection{Evaluation Metrics and Results}
Performance is evaluated using standard trajectory prediction metrics, including minimum Average Displacement Error (minADE), minimum Final Displacement Error (minFDE), and Miss Rate (MR), computed over the top-$K$ predicted trajectories.
\begin{table*}[t]
\caption{Trajectory prediction performance comparison on Nuscenes dataset. Metrics include $\operatorname{minADE}$ and $\operatorname{minFDE}$ for top-1 and top-10 predicted trajectories.}
\label{table::CAGMMresults}
\begin{adjustbox}{width=.99\linewidth ,center}

\begin{tabular}{lcccccc}
\hline
\textbf{Method} & \textbf{Model Type} & \textbf{Modes} & $\mathbf{minADE}_{1} \downarrow$ & $\mathbf{minFDE}_{1} \downarrow$ & $\mathbf{minADE}_{10} \downarrow$ & $\mathbf{minFDE}_{10} \downarrow$ \\
\hline
Constant Speed \& Heading & Baseline       & 1       & 4.61 & 11.21 & N/A  & N/A  \\
MTP (Single Mode) \cite{cui2019multimodal}        & Raster-based   & 1       & 4.17 & 9.37 & N/A  & N/A  \\
MTP (Multi-Modal) \cite{cui2019multimodal}        & Raster-based   & 16      & 4.55 & 9.58 & 3.25 & - \\
CoverNet, $\epsilon = 3$ \cite{phan2020covernet} & Raster-based   & 357    & 3.90 & 9.65 & 1.57 & 5.31 \\
MultiPath \cite{chai2019multipath}                & Raster-based   & 64       & 4.05 & 9.47 & 3.33 & - \\
CASPFormer \cite{wu2024caspformer}               & Raster-based   & 5       & - & 6.702 & 1.15 & - \\
AgentFormer \cite{yuan2021agentformer}              & Both           & -       & - & - & 1.45 & 2.86 \\
SemanticFormerR \cite{sun2024semanticformerholisticsemantictraffic}          & Graph-based    & 10      & - & 6.27 & 0.87 & - \\
LaPred++ \cite{kim2021lapred}                  & Graph-based    & 10       & 3.51 & 7.58 & 1.12 & 2.39 \\
PGP \cite{pmlr-v164-deo22a}                      & Graph-based    & 10       & - & 7.17 & 1.00 & 2.80 \\
LAformer \cite{liu2024laformer}                  & Graph-based    & 10      & - & - & 0.93 & 
1.50 \\
DGCN                      & Graph-based    & 10      & - & 3.62 & 0.97 & - \\
CA-GMM (Ours)  & Raster-based   & $33 \times 10$ & \textbf{4.01} & \textbf{8.93} & \textbf{2.13} & \textbf{4.91} \\
CA-GMM (Ours)  & Raster-based   & $15 \times 6$ & \textbf{3.89} & \textbf{8.11} & \textbf{2.05} & \textbf{4.70} \\
CAA-GMM (Ours)  & Raster-based   & $33 \times 10$ & \textbf{2.15} & \textbf{5.68} & \textbf{1.01} & \textbf{2.11} \\
CAA-GMM (Ours)  & Raster-based   & $15 \times 6$ & \textbf{1.85} & \textbf{5.29} & \textbf{0.89} & \textbf{1.81} \\
\hline
\end{tabular}

\end{adjustbox}
\end{table*}

\begin{equation}
\label{metrics}
\begin{gathered}
\mathrm{minADE}_K =
\min_{k \le K}\frac{1}{T}\sum_{t=1}^{T}
\|\hat{y}_k^{(t)} - y^{(t)}\|_2, \\[2pt]
\mathrm{minFDE}_K =
\min_{k \le K}
\|\hat{y}_k^{(T)} - y^{(T)}\|_2, \\[2pt]
\mathrm{MR}_K =
\frac{1}{N}\sum_{i=1}^{N}
\mathbb{I}\!\left[
\min_{k \le K}
\|\hat{y}_{i,k}^{(T)} - y_i^{(T)}\|_2 > \delta
\right].
\end{gathered}
\end{equation}

All networks are trained end-to-end using the loss defined in Section~\ref{sec::training}. For nuScenes, 2\,s of agent history is used, while for Argoverse~2 we employ 5\,s (benchmark evaluation) and 3\,s (ablation) histories. In ablation experiments, the availability of neighboring agents’ histories is selectively removed to simulate communication and observation losses.

Tables~\ref{table::CAGMMresults} and~\ref{tab:argoverse2_sota} report quantitative results on nuScenes and Argoverse~2 with a 6\,s prediction horizon. On nuScenes, CA-GMM consistently outperforms raster-based baselines such as MTP and CoverNet, demonstrating the benefit of explicit Gaussian-based probabilistic inference for modeling multimodal futures. 

The proposed CAA-GMM further improves performance by incorporating attention-based interaction modeling, achieving competitive or superior accuracy compared to recent attention-based raster models such as CASPFormer and AgentFormer. Notably, CAA-GMM approaches the performance of advanced graph-based methods (e.g., SemanticFormerR and LAformer) without requiring explicit relational graph construction or costly preprocessing.
\begin{table}[t]
\centering
\caption{Comparison with state-of-the-art methods on the Argoverse~2 validation set.}
\label{tab:argoverse2_sota}
\resizebox{\columnwidth}{!}{
\begin{tabular}{lccccc}
\toprule
\textbf{Method} & \textbf{minADE$_1$} & \textbf{minFDE$_1$} &
\textbf{minADE$_6$} & \textbf{minFDE$_6$} & \textbf{MR$_6$} \\
\midrule
QML w/ ensemble~\cite{qml2023} & 1.84 & 4.98 & 0.69 & 1.39 & 0.19 \\
BANet w/ ensemble~\cite{banet2022} & 1.79 & 4.61 & 0.71 & 1.36 & 0.19 \\
HPTR~\cite{hptr2023} & 1.84 & 4.61 & 0.73 & 1.43 & 0.19 \\
SIMPL~\cite{simpl2024ral} & 2.03 & 5.50 & 0.72 & 1.43 & 0.19 \\
FR~\cite{fr2023} & 2.37 & 5.93 & 0.89 & 1.81 & 0.29 \\
Forecast-MAE~\cite{cheng2023forecastmae} & 1.84 & 4.60 & 0.73 & 1.43 & 0.18 \\
TrajAngleNet~\cite{bharilya2024trajanglenet} & 1.75 & 4.40 & 0.70 & 1.36 & 0.17 \\
\textbf{CAA-GMM (Ours)} & \textbf{1.89} & \textbf{4.71} & \textbf{0.75} & \textbf{1.39} & \textbf{0.18} \\
\bottomrule
\end{tabular}}
\end{table}
On Argoverse~2, CAA-GMM exhibits strong generalization across complex traffic scenarios and long-horizon prediction, achieving performance comparable to high-capacity graph- and query-based approaches such as TrajAngleNet, QML, and BANet. These results indicate that combining probabilistic Gaussian inference with lightweight attention mechanisms yields a favorable balance between predictive accuracy, computational efficiency, and scalability.

Finally, ablation results reveal a trade-off between the number of Gaussian components and generalization. While increasing the number of modes improves representational capacity, excessively large mixtures can degrade performance due to overfitting and mode fragmentation. This highlights the importance of selecting an appropriate number of Gaussian modes to balance prediction diversity and robustness.
\subsection{Ablation Under Communication and Observation Loss}
\label{sec:ablation}

We evaluate the robustness of CAA-GMM under realistic communication and perception degradations on the Argoverse~2 validation set using a 3\,s history to predict a 6\,s future trajectory. All vehicles are assumed connected via C-V2X PC5 sidelink. During the history window ($t\in[2,5]$\,s), neighboring vehicles broadcast their current states and self-predicted trajectories; at $t=5$\,s, the focal vehicle assembles the most recent available states and forecasts its trajectory over $(5,11]$\,s.

\subsubsection{Communication Loss Model}
Following empirical LTE-V2X Mode~4 studies~\cite{bazzi2020analytical,bazzi2021empirical,molina2020analytical,3gppTR36885}, packet success is modeled as a Bernoulli process dependent on distance $d$, traffic density $\rho$, message rate $\lambda$, and packet size $S$. Packet airtime and channel busy ratio (CBR) are
\begin{equation}
T_{\mathrm{pkt}}=\frac{8S}{R_b}, \qquad
\mathrm{CBR}=1-\exp(-\rho\,\lambda\,T_{\mathrm{pkt}}\,L_{\mathrm{sense}}),
\end{equation}
with PHY rate $R_b=6$\,Mb/s and carrier-sense range $L_{\mathrm{sense}}=300$\,m. Conditioned on line-of-sight (LOS) or non-line-of-sight (NLOS),
\begin{equation}
P_{\mathrm{succ}\mid s}
=\sigma\!\big(
- a^{(s)}_0 - a^{(s)}_1 d - a^{(s)}_2 \mathrm{CBR} - a^{(s)}_3 T_{\mathrm{pkt}}
\big),
\end{equation}
where $(a_0,a_1,a_2,a_3)=(-3.0,0.014,4.0,1200)$ for LOS and $(-2.3,0.020,4.5,1500)$ for NLOS. The marginal success probability mixes LOS/NLOS using the distance-dependent LOS probability from 3GPP~TR~38.901~\cite{3gppTR38901}:
\begin{equation}
P_{\mathrm{succ}} =
p_{\mathrm{LOS}} P_{\mathrm{succ}\mid \mathrm{LOS}} +
(1-p_{\mathrm{LOS}}) P_{\mathrm{succ}\mid \mathrm{NLOS}}.
\end{equation}

\subsubsection{Observation Loss Model}
Detection of pedestrian and cyclist agents is modeled as a function of distance $d$ and occlusion ratio $r_o$, following probabilistic visibility formulations~\cite{ibrahim2022visibility,koc2021emergence,safe_occlusion_driving}:
\begin{equation}
P_{\mathrm{det}}(d,r_o)=
\frac{1}{1+\exp\!\big(
\theta_0 + \theta_1 \ln d + \theta_2 d + \theta_3 r_o + \theta_4 d\,r_o
\big)},
\end{equation}
with $(\theta_0,\theta_1,\theta_2,\theta_3,\theta_4)=(-6.3,1.25,0.030,1.6,0.06)$.

\subsubsection{Ablation Results}
Lost communication packets are replaced by the most recent cached self-predicted state, while missed detections are linearly interpolated; agents with fewer than three detections over the 3\,s history are discarded.

CAA-GMM is evaluated on three traffic-load subsets of the Argoverse~2 validation set (low, medium, congested), constructed to preserve the original agent-count distribution. For each subset, performance is reported under stochastic communication and detection losses (“Prob.”) and under perfect communication and perception (“Perf.”) on the same scenes.

\begin{table}[t]
\centering
\caption{CAA-GMM performance on Argoverse~2 under varying traffic-load conditions.}
\label{tab:caagmm_ablation_min6}
\setlength{\tabcolsep}{2.8pt}
\renewcommand{\arraystretch}{1.1}
\footnotesize
\resizebox{\columnwidth}{!}{%
\begin{tabular}{lcccccc}
\toprule
\multirow{2}{*}{\textbf{Load}} &
\multirow{2}{*}{\textbf{\#Samples}} &
\multirow{2}{*}{\textbf{Agents}} &
\multicolumn{2}{c}{\textbf{minADE$_6$} $\downarrow$} &
\multicolumn{2}{c}{\textbf{minFDE$_6$} $\downarrow$} \\
\cmidrule(lr){4-5}\cmidrule(lr){6-7}
 & & & \textbf{Prob.} & \textbf{Perf.} & \textbf{Prob.} & \textbf{Perf.} \\
\midrule
Low       & 108  & $<\!20$   & 0.91 & 0.89 & 1.75 & 1.67 \\
Medium    & 547  & 20--40    & 0.89 & 0.81 & 1.69 & 1.52 \\
Congested & 1345 & $>\!40$   & 1.07 & 0.83 & 1.96 & 1.53 \\
\bottomrule
\end{tabular}%
}
\vspace{2pt}
\footnotesize
\emph{Notes.} “Prob.” denotes stochastic communication and detection loss; “Perf.” assumes perfect communication and perception on the same subset.
\end{table}

As shown in Table~\ref{tab:caagmm_ablation_min6}, CAA-GMM exhibits minimal degradation under low and medium traffic loads and only a moderate increase in error under congested conditions, demonstrating robustness to realistic joint communication and observation impairments.
\section{Concluding Remarks}
\label{sec:chap3_concl}
This work introduced CA-GMM and its attention-augmented extension, CAA-GMM, for probabilistic trajectory prediction in cooperative driving. Experiments on nuScenes and Argoverse~2 show that combining Gaussian mixture inference with attention-based interaction modeling yields accurate and interpretable multimodal predictions, achieving performance comparable to state-of-the-art graph-based methods without explicit relational graphs or costly preprocessing. Robustness evaluations under realistic communication and perception losses further demonstrate that CAA-GMM maintains stable prediction accuracy under imperfect connectivity and partial observability.

\bibliographystyle{IEEEtran}\small
\bibliography{references}

\end{document}